# State Classification of Cooking Objects Using a VGG CNN

Kyle Mott

*Abstract*—In machine learning, it is very important for a robot to know the state of an object and recognize particular desired states. This is an image classification problem that can be solved using a convolutional neural network. In this paper, we will discuss the use of a VGG convolutional neural network to recognize those states of cooking objects. We will discuss the uses of activation functions, optimizers, data augmentation, layer additions, and other different versions of architectures. The results of this paper will be used to identify alternatives to the VGG convolutional neural network to improve accuracy.

## I. INTRODUCTION

Robotics is increasingly becoming more and more applicable to everyday life's tasks. Robots have been able to replace humans in completing day to day tasks in the kitchen. For a robot to perform meal preparation tasks, it needs to be able to recognize different cooking states of objects and perform those actions. A robot being able to perform object state recognition can be very useful towards identifying correct states concurrently throughout cooking processes. While performing those actions, the robot must identify the end of one current state to another in order to satisfy completion of a particular task. For example when peeling a carrot, the robot will use object state classification to recognize the state of the whole carrot, then using grasping techniques to pick up the carrot and tool to perform the peeling task. Both object state classification and grasp planning are important for a robot to complete day to day tasks in the kitchen.

Various techniques of robot grasp planning [1], [2], and [3] have been achieved over the years. Grasp planning has been achieved using probabilistic inference [1]. Modeling and probabilistic framework [1] of robot-grasping tasks have been used to provide insights for future research to grasp planning for robots. Power and dexterity manipulation tasks [2] in grasp planning are important in various tools the robot will grasp in the kitchen environment. A learning approach was used for grasp planning on unseen objects [2] encountered where precision and power for a given object are needed. Even robot learning of grasping unknown objects from human demonstration [3] have been proposed. Robots have learned from a video representation of appropriate grasping techniques of different directions of the target object [3].

Object image classification has been a vital component for robotics for various applications. Object image classification typically determines grasp planning in robotic applications. There have been different methods for identifying objects. As spatial resolution increases, there is more variance of individual pixels that are in the same class [4]. Because of this, object classification are more effective than pixel methods [4].

Meal preparation tasks will vary by portion and object state for various stages of cooking depending on the meal that will be made. For example when preparing carrots, the carrots change various states (e.g. whole, peeled, julienne, cooked, ect.) throughout the cooking process. The robot will need to perceive and identify the different states of the carrot at each stage. Only once the robot has clearly identified the object's completed state (all of the object) it can guarantee that the object is no longer in the previous state.

This report focuses on using deep convolutional neural networks with a VGG base model to achieve image classification between 11 different object states (creamy paste, diced, floured, grated, juiced, julienne, mixed, other, peeled, sliced, and whole). The training dataset consists of 6348 images and the validation dataset consists of 1377 images of cooking objects in the previous states. The goal is to grasp insight on properly detecting the cooking states based on the unseen data. We will indulge into architectural design of convolutional neural networks and different structures to see changes in accuracy.

## II. DATA COLLECTION AND PREPROCESSING

### A. Dataset

The dataset for this project includes 17 different cooking objects (chicken/turkey, beef/pork, tomato, onion, bread, pepper, cheese, strawberry, ect.) with 11 different object states (creamy paste, diced, floured, grated, juiced, julienne, mixed, other, peeled, sliced, and whole) [5]. The total dataset contains 9309 total images [6], of which 6348 training images, 1377 validation images, and the rest test images. The dataset can be seen here [6], dataset version 1.2.

### B. Preprocessing and Data Augmentation

The raw images of the dataset can vary and are inconsistent. For instance, each image can be zoomed out, have unnecessary pixel data, or contain noise. A type of preprocessing and noise removal technique with MRI images [7] was used using different types of filters to get accurate observations.

First, the data is partitioned into 3 datasets: training, validation, and testing. Out of 9309 total images, roughly 68.2% was dedicated toward training, 14.8% was dedicated toward validation, and 17.0% toward testing. The images are organized by folders containing the appropriate state, but the images are shuffled and random.

The input images from the dataset vary in size and shape. The larger the image size the more noise that's generated which takes more computation and epochs. The smaller the image, the more vital information about the actual image is lost which will cause problems during classification. A good explanation of this theory can be found in this video [8]. We eventually chose an image size of 150 by 150. This could contribute to an increase in accuracy if we had chosen a higher image size and then ultimately would have to increase the computation and number of epochs.

Data augmentation is an effective tool that can be used with low amount of images in our dataset. With data augmentation we can help reduce overfitting by increasing the size of the dataset through the use of rescaling, reflection, rotation, etc. The ImageDataGenerator factors used for data augmentation and normalization for this project are listed in the following Table 1. The same data augmentation factors were used for all data partitions.

Table 1: Data Augmentation Factors

| Type | Factor |
| --- | --- |
| Shear Range | 0.2 |
| Zoom Range | 0.2 |
| Horizontal Flip | True |
| Fill Mode | Nearest |
| Rescale | 1/225 |
| Height Shift Range | 0.2 |
| Width Shift Range | 0.2 |
| Rotation Range | 40 |

Without data augmentation the model was overfitting with the low and amount of dataset images. With the current image size and data augmentation factors, up to around 50 epochs were needed to reach a stable accuracy.

III. METHODOLOGY

A modified version of the VGG19 network was used to classify the cooking objects for this dataset. At first, all the base layers of the VGG19 network were used and layers were then frozen and convolution was added to increase accuracy. The amount of layers added, amount of base layers frozen, and amount of filter sizes were all attempted to increase results. Many of the lower base layers of the VGG19 model weren't frozen because those layers would only grab basic features of the images while the outer most layers would start to grab more intricate details separating themselves from other states. Because of this, only the top few layers were frozen then new layers were added on.

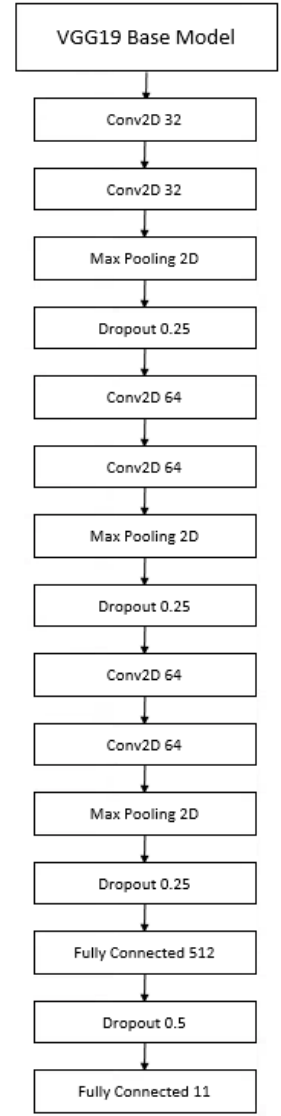

*Figure 1: Modified VGG19 Architecture*

In our model, we added two convolutional layers of 32, max pooling, then dropout. I repeated this process two more times but convolutional layers of 64 instead and ended with a dense layer of 512, dropout of 0.5 and output dense layer of 11 cooking states. A method of adding more convolutional layers, max pooling, and dropout was tried; but, resulted in reducing the overall accuracy of the network. The result of this stems from the VGG19 base model, which become optimal around the 19th layer. Since a few layers were frozen, then new ones added, stemming over the 19th layer mark wouldn't add much optimization to the network. The original weights of the base model used were ImageNet [9].

Dropout of 0.25 for the individual layers and a dropout of 0.5 for the final dense layer was chosen for the final model. Increasing and decreasing the dropout per individual layer was tried. It was found that the decrease/removal of dropout in the layers increased the overall accuracy, but suffered from overfitting issues as a result [10]. Because of this, dropout was kept in the final model.

Many different optimizers were tried on this model. Adagrad was first used with the theory that the learning rate did not need to be changed very much [11]. The learning rate of Adagrad is always decreasing so the learning rate was left at a value of 0.001 and not changed. This was used to focus changing on layers and regularization in attempts to increase accuracy. SGD was chosen after most of the layers and regularization was setup in the modified architecture. One disadvantage of SGD is that it can converge very slowly toward parameter values. This was a problem because it was already taking extremely long to run this neural network on large amount of epochs. Also, it was difficult to settle on an appropriate learning rate for SGD. The optimizers tested for this modified architecture were Adagrad, Adam, Adamax, Nadam, RMSprop, and SGD. Ultimately, Adam was the final choice for this modified architecture due to its well performance and convergence time.

## IV. EVALUATION AND RESULTS

In order to optimally obtain graphical results of loss and accuracy, a visualization tool called TensorBoard [12] was used. This tool is integrated by reading the TensorFlow event files of the summary data collected from each architecture tested. TensorBoard can plot the graphs of each loss and accuracy results from multiple event files and compare the results. A smoothing definition of 0.5 was used for the plots shown.

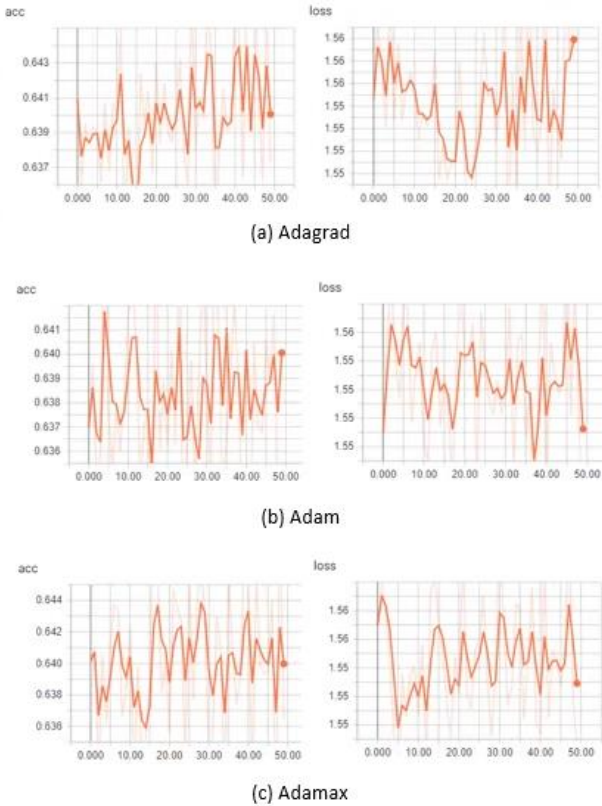

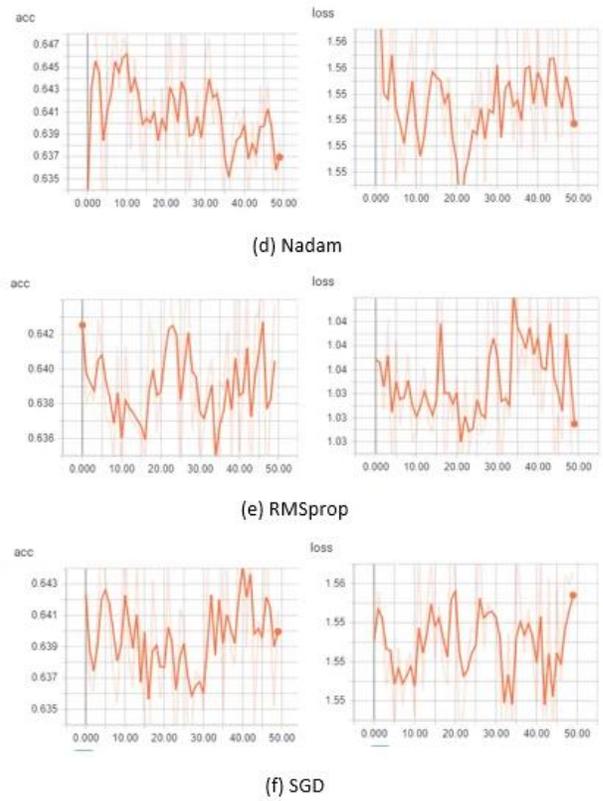

*Figure 2: Accuracy and Loss Results from Various Optimizers*

From the different optimizers tested, there wasn't drastic changes in accuracy from each optimizer while using the same weights. All of the optimizers hovered around 0.639 to 0.642 accuracy within 50 epochs. The Nadam (d) optimizer did however result in a lower loss than the other optimizers at around 1.03 while the other optimizers resulted in approximately 1.55 to 1.56 loss. Because of the overall insignificance of change between optimizers, it was decided to use Adam (b) as the optimizer for the final modified model while knowing that the convergence is reasonable and not much changed with the same weights used.

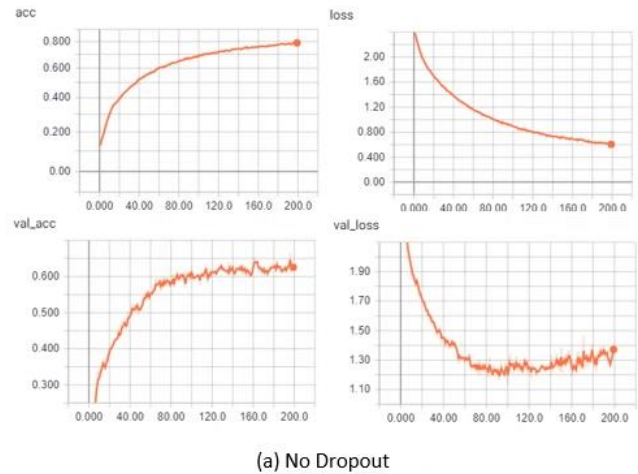

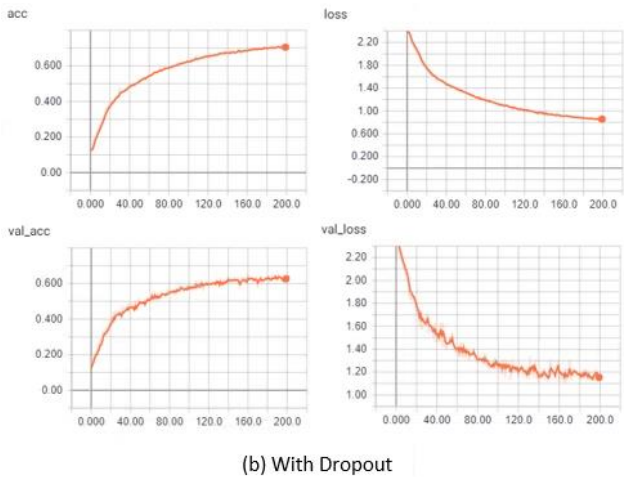

*Figure 3: Dropout Accuracy and Loss Results*

The regularization method used in this modified VGG network was dropout [10]. With using dropout, it was expected to reduce overfitting. Results were taken with a dropout of 0.25 occurring at each individual hidden layer (a) and results without dropout at each individual hidden layer (b). From (a), there was a resultant accuracy of 78.7%, with a validation accuracy of 62.9%. While (b), there was a resultant accuracy of 69.8%, with a validation accuracy of 64.0%. While removing the dropout increased the overall accuracy, overfitting issues started to form confirming preliminary predictions. For this reason, dropout after each individual pooling layers was found in the final modified architecture.

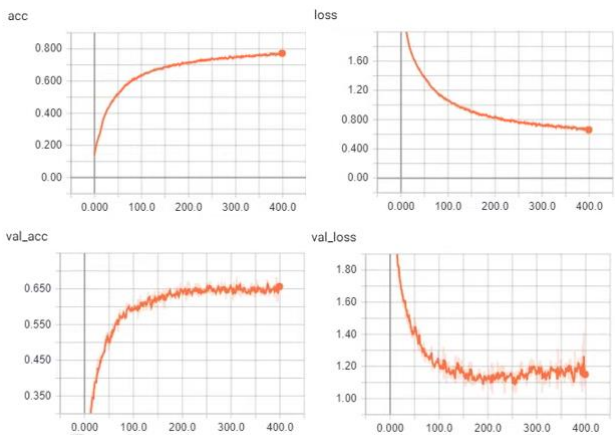

*Figure 4: 400 Epochs Accuracy and Loss Results*

From Fig. 4 we can see that even after 400 epochs compared to 200 epochs (Fig. 3 (a)), the accuracy does increase slightly to around 78% from 69.8% while validation accuracy remains around the same. This shows that after 200 epochs overtraining becomes an issue and the model starts to over-fit to the training dataset.

After the best performing model was chosen to test unseen data, it obtained an accuracy of 37.7%. A factor that could have contributed to this low accuracy is that the data was over-fit and not generalizing. VGG19 is a base model that performs very well around 19 layers and because this modified model has more than that could be another factor to contribute toward low accuracy. With an image size of 150 by 150, increasing the image size would take much more convolution but inevitably result in a higher accuracy. Data augmentation factors could have been changed to increase the dataset to help reduce over-fitting issues.

## V. CONCLUSION

Image classification is just one of the main challenges faced in robotics. To deal with this challenge, a modified VGG19 convolutional neural network was chosen to perform training and test on unseen data. The model was designed to recognize the object of 11 different states (creamy paste, diced, floured, grated, juiced, julienne, mixed, other, peeled, sliced, and whole). With a total of 9309 images in the dataset, the modified network obtained an accuracy on unseen test data of 37.7%. This study was challenging due to the modified network chosen being easily susceptible to over-fitting. Something that could be changed to increase the accuracy are modification of the convolutional layers. Increasing the input image size and data augmentation factors would contribute to increase accuracy and reduction of over-fitting. For the future, more analysis will need to be made to reduce over-fitting and other convolutional neural network architectures can be explored for image classification.